\documentclass[11pt]{article}
\usepackage[margin=1in]{geometry}
\usepackage{amsmath,amssymb,booktabs,graphicx,hyperref,xcolor}
\usepackage{xurl}  %
\hypersetup{colorlinks=true,linkcolor=blue,citecolor=blue,urlcolor=blue}
\usepackage{authblk}
\usepackage[numbers]{natbib}

\title{Behaviour Is an Incomplete Measure of Reasoning Development:\\
\large Cross-surface pre-arrival accessibility and the limits of developmental inference
in a recurrent-depth reasoner}
\author[1]{Simon Lam-Muir}
\affil[1]{Prime Calibre, Australia \quad \texttt{research@primecalibre.com} \quad ORCID 0009-0001-2442-4479}
\date{August 2026}

\begin{document}
\maketitle

\begin{abstract}
Capability development is routinely inferred from behavioural thresholds, from final checkpoints, or
from what a decoder can read out of a hidden state. These quantities need not identify the same
event. We study a 30M-parameter recurrent-depth relational reasoner in a closed, oracle-defined
world, using dense behavioural trajectories, two training surfaces, preregistered pre-arrival
hidden-state probes, prospectively checked evaluability, and explicit untrained and negative
controls, holding the training-time and inference-time axes separate throughout. Behaviour first:
under one frozen acquisition criterion, three-hop competence cost $70$ logical epochs on the symbolic
surface and $13{,}055$ on the verbal surface --- a derived contrast of $186.5\times$ --- after which
verbal four-hop competence cleared in $8$ logical epochs. Across the $13{,}055$-epoch grind,
four-hop held-out behaviour never exceeded $3/40$ and ended at $0/40$. Internal measurement next:
on the verbal surface a linear probe recovered future-answer identity before behavioural arrival at
$0.056159$ against uniform chance $0.025$, an untrained control of $0.024758$ and a population
frequency baseline of $0.048309$ ($p = 0.012987$; $16/40$ answer classes contributing). Analogous pre-arrival accessibility survived
the surface change, reaching $0.1020$ against a zero-step control of $0.0460$
($p = 0.000999$) at the upstream structural position and $0.0618$ at the readout comparator
($p = 0.004$), with $21/40$ classes contributing. Finally, the natural attempt to track that
accessibility across training was not cleanly evaluable: probe eligibility is defined by behavioural
arrival, so the measured population changes with the measurand. Behavioural competence, internal
accessibility, and training-time development are distinct observables, and neither behaviour nor
decoder accessibility identifies the computation training acquired; causal intervention is the
necessary next step.
\end{abstract}

\section{Introduction}
\label{sec:intro}

Three measurements are commonly used to say when a model acquired a capability. The first is
behavioural: a threshold on held-out accuracy, crossed at some point in training. The second is
internal: a decoder or probe recovers some property from hidden state, and the property is said to
have appeared when the probe succeeds. The third is retrospective: a final checkpoint is analysed and
the resulting mechanism is described as what training produced.

These three need not agree, and where they disagree, the disagreement itself does not tell us which
one corresponds to the computation the model actually acquired. That is the measurement problem this
paper addresses.

We study it in a setting chosen so that the ground truth is mechanically available: a closed
relational world with oracle-defined answers, a recurrent-depth
reasoner~\citep{geiping2025recurrentdepth} trained by a hop-depth curriculum, and two training
surfaces that express the same underlying relational task family differently. The closed world lets
us check item novelty, path overlap and answer-class coverage exactly, rather than estimating them.

\paragraph{Contributions.}
First, \emph{behavioural development has hidden structure}. Holding the world, architecture, curriculum, optimiser configuration and acquisition criterion
fixed while changing the bundled training surface produces radically different acquisition costs
and different developmental shapes --- long grinds on one surface, rapid races on the other. A single
competence threshold collapses that structure to a point.

Second, \emph{pre-arrival answer accessibility is not confined to one surface}. Weak, heterogeneous
information about the future answer is linearly accessible from hidden state before the model
behaviourally produces that answer, on the verbal surface, and this survives a change of surface,
appearing at two separately preregistered positions on the symbolic surface.

Third, \emph{richer observation still has identification limits}. The obvious next measurement ---
tracking pre-arrival accessibility across training --- cannot be validly estimated here, because the
population eligible for the probe is defined by the behavioural event being studied. Together with a
quarantined execution and an earlier validation branch that correctly returned no result, this shows
that an instrument's refusal to emit an unsupported number is itself informative.

This paper does not identify the acquired computation. It establishes why behavioural and
accessibility measurements alone are insufficient to do so, and why the next step must be causal.

\section{Experimental setting and measurement framework}
\label{sec:setting}

\subsection{An oracle-defined relational world}
The world contains a fixed set of entities and relations. A $k$-hop query names a starting entity and
a chain of $k$ relations; its answer is the entity reached by following that chain. Because the world
is closed and generated from a fixed seed, every answer is mechanically known, every intermediate
along a path is definable, and the novelty of any evaluation item with respect to training can be
checked exactly rather than estimated. Held-out batteries are constructed per depth with verified
zero overlap against training items.

This control is the reason for the setting, and also its principal limitation: results here concern
this world and this substrate class. Nothing in what follows should be read as automatically
transferring to models trained on natural language at scale.

\subsection{Model and curriculum}
The model is a 30M-parameter recurrent-depth transformer: a 768-dimensional state and a 4-layer block
that is applied repeatedly within a single forward pass, so that computational depth at inference is
decoupled from parameter count~\citep{geiping2025recurrentdepth}. Training proceeds by curriculum
over hop depth. A stage $k$ trains on all items of depth at most $k$ and is evaluated on a held-out
$k$-hop battery; the stage clears when held-out accuracy strictly exceeds $0.95$, a criterion frozen
before any of the results reported here.

Physical reuse of the recurrent block is an architectural fact. It is not evidence that the model
reuses a semantic operator across depths, and we do not treat it as such.

\subsection{Two surfaces}
The same relational task family is presented through two surfaces. The symbolic surface names
entities and relations by opaque tokens; the verbal surface expresses the same queries as English-like
sentences with function words and nested genitive structure. The world, the curriculum, the
architecture, the optimiser configuration and the acquisition criterion are held fixed across the two.

We are careful about a phrase that would be convenient and wrong. These are not the same model
evaluated twice; they are separate trained realizations under a matched architecture, world and
curriculum. Statements below compare realizations, not a single model's two behaviours.

\subsection{Developmental coordinates}
Acquisition costs are reported in \emph{logical epochs} under one convention, fixed in advance:
milestone checkpoints are named by a filename epoch, and the in-stage cost of stage $k$ is the
difference between the clearing epochs of stages $k$ and $k-1$. Distinct epoch namespaces exist in the
training system --- the filename coordinate, an internal payload field offset by one, and a
frame coordinate used by the developmental camera --- and every quantity in this paper is reported in
a named namespace rather than inferred from filename arithmetic alone.

Where a stage's lower boundary is not established by a surviving predecessor checkpoint, we report
the stage's clearing epoch but \emph{not} an in-stage cost. This applies to $k=2$ on both surfaces:
no predecessor clearance checkpoint exists, and epoch $0$ is not assumed. Those cells are recorded as
not reportable rather than imputed.

\subsection{Two axes that must not be merged}
Two different clocks run in this system. \emph{Training time} indexes parameter updates across
checkpoints. \emph{Inference time} indexes recurrent iterations inside a single forward pass. A
statement about when information becomes available inside one forward pass is not a statement about
when training produced it, and the reverse also holds. Keeping these separate is load-bearing for
everything in Sections~\ref{sec:act2} and~\ref{sec:act3}.

\subsection{Probe protocol and controls}
\label{sec:probes}
Probes are linear classifiers over hidden state at registered, event-relative positions, fitted
independently per layer across four layers and scored on held-out folds under a fold map frozen
before any activation was captured. Each probe result is reported against three references
simultaneously:

\begin{itemize}
\item \textbf{Uniform chance}, $1/40 = 0.025$ for the 40-answer-class universe;
\item \textbf{A population frequency baseline}, computed on the same rows, which absorbs any class
      imbalance the probe could otherwise exploit;
\item \textbf{An untrained zero-step control}, a model reconstructed deterministically at
      initialisation under the same procedure, which absorbs anything readable from architecture and
      inputs alone.
\end{itemize}

Significance is assessed by item-level restricted permutation with $B = 1000$ replicates; we report
the null mean, spread and maximum, not only the $p$-value. Class-level heterogeneity is reported
beside every pooled figure.

The zero-step control carries a classification that travels with every result derived from it:
it is a deterministic reconstruction from recovered historical provenance under a prospectively
frozen procedure, and its byte-identity with the original historical initialisation is unverified.

\subsection{Evaluability is checked before inference}
\label{sec:evaluability}
One design principle governs the internal measurements: \emph{question, then minimum sufficient
statistic, then a mechanical proof that the population supports that statistic, and only then the
smallest freeze that fixes it}. A planned analysis is not run merely because the activations exist.
The population must be shown, mechanically and in advance, to support the estimand.
Section~\ref{sec:act3} is what happens when it does not.

\section{Behaviour has hidden structure}
\label{sec:act1}

\subsection{A surface-dependent acquisition staircase}
Table~\ref{tab:staircase} reports the early acquisition staircase numerically, while
Figure~\ref{fig:behaviour}A shows the longer symbolic trajectory through $k=20$ alongside the available
verbal stages, all under the same frozen criterion. Three-hop competence cost $13{,}055$ logical epochs on the
verbal surface against $70$ on the symbolic surface, a derived contrast of $186.5\times$. The symbolic
staircase then proceeds in small steps --- $6$, $1{,}099$, $2$, $1$, $1$, $1$ epochs at successive
depths --- while the verbal trajectory is dominated by the $13{,}055$-epoch $k=3$ grind, followed by
an eight-epoch $k=4$ race and a second long, right-censored $k=5$ stage that had accumulated at least
$6{,}400$ logical epochs at the manuscript snapshot.

The result is not the ratio. It is that two realizations of the same relational task family, matched
in architecture, world, curriculum and criterion, have developmental trajectories of entirely
different shape. A single threshold reports one number and discards that shape.

\subsection{A long grind followed by an eight-epoch race}
After $13{,}055$ epochs at $k=3$, the verbal realization cleared $k=4$ in $8$ logical epochs.

We state that as a sequence and stop there. The grind preceded the race; behaviour alone does not
tell us why. In particular, this observation does not establish that the grind purchased machinery
later reused at $k=4$, and it does not establish that four-hop capability was present and unexpressed.
Both remain open, and Section~\ref{sec:act3} explains why the measurement that would most directly
address them is not available here.

\subsection{Later-depth behaviour during the grind}
\begin{figure}[t]\centering
\includegraphics[width=\textwidth]{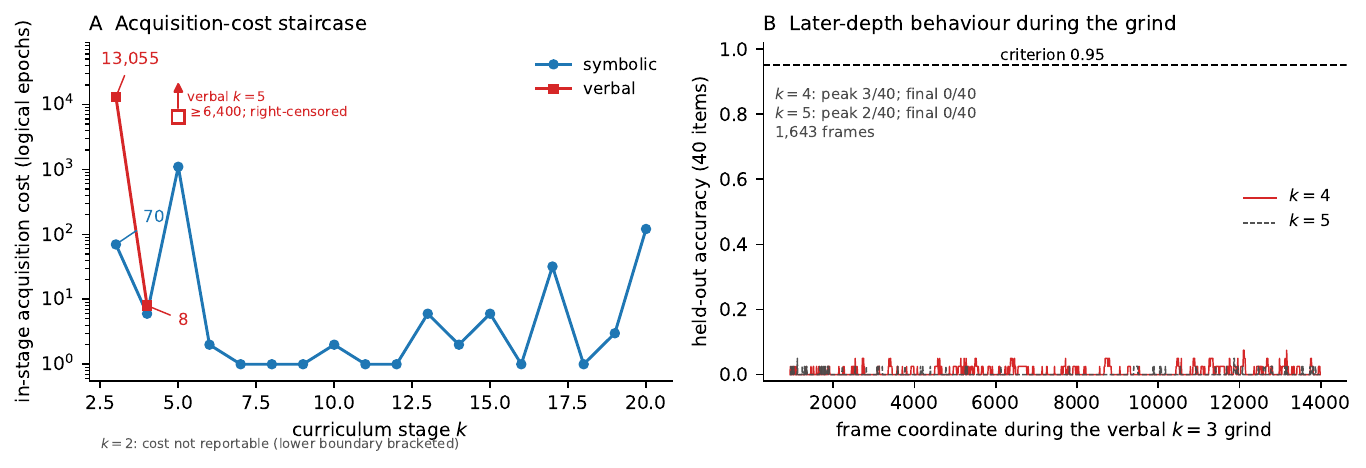}
\caption{\textbf{Behavioural development.} \textbf{(A)} In-stage acquisition cost by curriculum stage
for both realizations, logarithmic scale. Verbal $k=3$ costs $13{,}055$ logical epochs against
symbolic $70$ at the same depth, a derived contrast of $186.5\times$; verbal $k=4$ then costs $8$.
Verbal $k=5$ is plotted at its manuscript-snapshot lower bound of ${\geq}6{,}400$ logical epochs using an
open marker and upward arrow to denote right-censoring; this is not a completed stage cost. The $k=2$
costs are not plotted because their lower boundaries are bracketed and exact in-stage costs are not
reportable. The symbolic series
is shown through $k=20$; Table~\ref{tab:staircase} prints the $k\leq 10$ subset for compactness, and both
are generated from the same authoritative staircase source.
\textbf{(B)} Held-out four- and five-hop accuracy across $1{,}643$ frames of the verbal $k=3$ grind,
against the $0.95$ criterion. Four-hop peaks at $3/40$ and ends at $0/40$; five-hop peaks at $2/40$ and ends at $0/40$.
Both are at floor scale throughout. Plotted directly from the registered per-frame series.
\textbf{Panels use different developmental coordinates:} panel A reports logical-epoch stage costs,
whereas panel B uses the registered frame coordinate; horizontal positions are not directly
comparable across panels.}
\label{fig:behaviour}
\end{figure}

Figure~\ref{fig:behaviour}B shows held-out four- and five-hop accuracy across $1{,}643$ frames of the
verbal $k=3$ grind. Four-hop accuracy took only the values $0$, $0.025$, $0.05$ and $0.075$; it was
exactly zero in $71.3\%$ of frames, peaked at $3/40$, and ended at $0/40$. Five-hop accuracy took only
$0$, $0.025$ and $0.05$, was zero in $93.3\%$ of frames, and also ended at $0$. The criterion is
$0.95$. Against that criterion both curves are at floor throughout.

\paragraph{A classifier that failed, reported beside the raw curve.}
A five-class developmental classifier was registered in advance to label these curves, and its labels
were emitted: \textsc{distributed-climb} for $k=4$ and \textsc{none-of-the-above} for $k=5$. Those
labels misdescribe the data, and the reason is instructive. The classifier's leading statistic is
maximum-based, so a single $3/40$ frame among $1{,}643$ is enough to defeat its \textsc{flat}
predicate; a concentration measure computed over roughly seventy isolated floor flickers then becomes
small, which selects \textsc{distributed-climb}. A pure-noise floor curve of this length will
reliably receive that label. The $k=5$ curve escaped only through an exact equality in a threshold
comparison --- two contributing items rather than three is the whole difference between the two
labels.

We report the registered labels because they were registered, and we rest the interpretation on the
raw values. The methodological point is general: extreme-value statistics applied to very long
developmental series are not robust, and a classifier validated on short series should not be
transported to long ones without re-registration.

\subsection{What a behavioural threshold conceals}
Surface representation changes the developmental trajectory profoundly. An eventual criterion
crossing says little about the route taken to it. And the absence of competence-scale later-depth
behaviour during a long grind does not establish the absence of internal development, because
behaviour measures expression, not availability. That is the motivation for measuring inside the
model --- and the subject of the next section.

\begin{table}[t]\centering\small
\caption{Acquisition-cost staircase, both realizations, one frozen convention
(strict $>0.95$ held-out accuracy; filename-epoch coordinate). In-stage cost is the difference
between successive clearing epochs. Both $k=2$ rows are \emph{not reportable}: no predecessor clearance checkpoint survives and epoch $0$
is not assumed. Symbolic rows through $k=10$ are shown here for tabular readability; Figure~\ref{fig:behaviour}A
plots the complete symbolic series through $k=20$ from the same authoritative staircase source. Verbal
$k=5$ is right-censored at the manuscript snapshot. Generated from \texttt{T1-acquisition-staircase.csv}.}
\label{tab:staircase}
\begin{tabular}{llrrl}\toprule
surface & stage $k$ & clearing epoch & in-stage cost & boundary status \\\midrule
symbolic & 2 & 2{,}808 & --- & bracketed lower bound \\
symbolic & 3 & 2{,}878 & 70 & exact \\
symbolic & 4 & 2{,}884 & 6 & exact \\
symbolic & 5 & 3{,}983 & 1{,}099 & exact \\
symbolic & 6 & 3{,}985 & 2 & exact \\
symbolic & 7 & 3{,}986 & 1 & exact \\
symbolic & 8 & 3{,}987 & 1 & exact \\
symbolic & 9 & 3{,}988 & 1 & exact \\
symbolic & 10 & 3{,}990 & 2 & exact \\
\midrule
verbal & 2 & 936 & --- & bracketed lower bound \\
verbal & 3 & 13{,}991 & \textbf{13{,}055} & exact \\
verbal & 4 & 13{,}999 & \textbf{8} & exact \\
verbal & 5 & --- & --- & right-censored at snapshot \\
\bottomrule\end{tabular}\end{table}

\section{Internal accessibility is a different observable}
\label{sec:act2}

\subsection{A local content-first observation during training}
In a registered assay over the developmental film, a content-sensitive differential reached $93.25$
against a parse-control baseline of $88.5$ and a confound boundary of $89.5$ fixed in advance ---
a clearance of $3.75$ --- at a frame where held-out novel two-hop accuracy was $0.025$.

The licensed reading is narrow. The content-sensitive differential exceeded both its parse control and
the preregistered confound threshold while novel behavioural accuracy remained near floor. The margin
over the boundary is modest and travels with the result. This is a \emph{local descriptive ordering
observation}: it is not an estimate of when any mechanism was acquired, and it is not the
developmental accessibility curve that Section~\ref{sec:act3} shows to be non-identifiable. It is also
a different observable class from the probe results below --- interior reads from a training film,
versus decoder probes on frozen checkpoints at registered event-relative positions --- and we do not
merge the two into a single notion of ``internal signal''.

\subsection{Verbal pre-arrival accessibility}
On the verbal surface, a linear probe was fitted at a registered position preceding the answer slot,
on items the model had not yet answered correctly, and scored on held-out folds. Its four-layer mean
held-out accuracy was $0.056159$, against uniform chance $0.025$, an untrained zero-step control of
$0.024758$, and a population frequency baseline of $0.048309$. Restricted permutation with $B = 1000$
gave $p = 0.012987$, with a null mean of $0.0285$ and a null maximum of $0.0658$. Sixteen of the forty
answer classes contributed any correct held-out prediction.

The reading is that future-answer identity is weakly but reproducibly linearly accessible before
behavioural arrival, on unseen item and path surfaces within the same frozen answer-class universe.

Four qualifications travel with it. The effect is weak in absolute terms. Class expression is strongly
heterogeneous. This is not a high-accuracy decoder and does not establish that a completed answer
representation exists. And accessibility is not use: a linear classifier can exploit distributed
partial information without that information being what the model's own computation consumes.

\subsection{Analogous accessibility survives a surface change}
The analogous measurement on the symbolic surface used two registered locations, fixed in advance: a
primary position, defined as the final observed query position immediately preceding the answer slot,
and a secondary readout comparator. At the primary position the trained probe reached $0.10201$
against a zero-step control of $0.04598$, with $p = 0.000999$ --- the observed value falls outside the
entire null distribution over $1000$ permutations, whose maximum was $0.06394$. At the secondary
position it reached $0.06178$ with $p = 0.003996$. Twenty-one of forty classes contributed at the
primary position and sixteen at the secondary.

At the primary position the zero-step control and the frequency baseline are numerically identical at
$0.04598$. This is a coincidence of this population, not an identity of the two quantities, and they
are reported as two separate references.

The two locations were registered separately, each with its own threshold: the secondary cannot
rescue the primary, and there is no omnibus criterion under which a positive at either location
counts.

Analogous pre-arrival answer accessibility therefore survives the surface change, at both registered
locations, with the upstream structural position carrying the stronger signal. Neither result
establishes causal use, and neither identifies the acquired computation.

\paragraph{On comparing the two surfaces.}
The symbolic primary sits $+5.60$ points above its frequency baseline where the verbal result sits
$+0.785$ above its own; the symbolic figure is $4.1\times$ uniform chance and $2.2\times$ its control.
These are descriptive contrasts. No registered cross-population effect-size comparison was performed,
and we do not claim the symbolic effect is statistically larger than the verbal one.

\subsection{Ignition as decisive expression, not first appearance}
Taken with earlier work on this substrate~\citep{lammuir2026ignition}, in which the resolution of an
answer at the vocabulary readout is a sharp interface event, the present results refine what that event marks. Ignition is not
the first appearance of answer-relevant information; it is the decisive readout commitment of
information that can already be weakly accessible internally --- and that refinement now holds on both
surfaces.

This synthesis relates two different variables and we mark the seam. The earlier readout result~\citep{lammuir2026ignition}
concerned composed intermediates exposed through the tied vocabulary readout; the probes here concern
final-answer identity present in hidden state. The relation is \emph{present} versus \emph{exposed},
not a two-channel comparison of one variable. We do not conclude that the model has already computed
the answer.

\begin{figure}[t]\centering
\includegraphics[width=\textwidth]{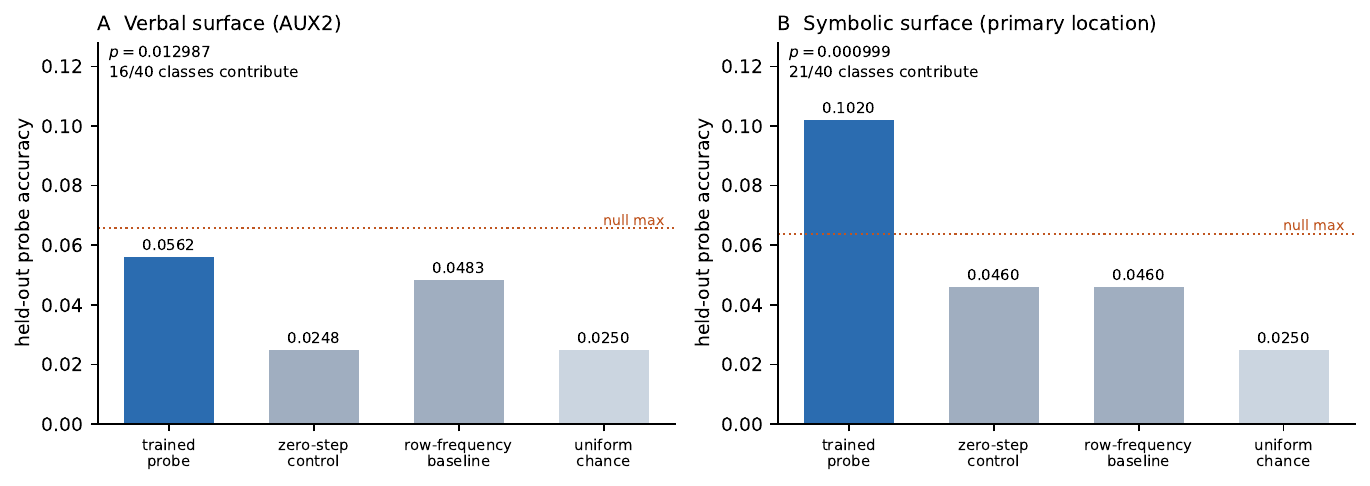}
\caption{\textbf{Pre-arrival answer accessibility on both surfaces.} Trained held-out probe accuracy
against all three registered references --- untrained zero-step control, population frequency
baseline, and uniform 40-way chance --- with the maximum of the $B=1000$ permutation null marked.
\textbf{(A)} Verbal surface. \textbf{(B)} Symbolic surface, primary position. Class heterogeneity is
printed with each panel. The two panels are shown together for readability; the verbal--symbolic
contrast is descriptive and no registered cross-population effect-size test was performed.}
\label{fig:accessibility}
\end{figure}

\section{Measurement has hard limits}
\label{sec:act3}

\subsection{A developmental curve that could not be estimated}
The natural next question is whether pre-arrival accessibility emerges or strengthens across a long
grind. The registered design was to fit the same probe at successive checkpoints and report the
resulting trajectory.

That estimand does not cleanly exist here, for a structural reason. Probe eligibility is defined by
behavioural arrival: an item is eligible only while the model has not yet answered it correctly. As
training proceeds, items improve and leave the eligible population --- and they leave \emph{because}
they improve. The population being measured therefore changes as a function of the measurand. We name
the failure mode \textbf{selection coupled to the measurand}.

The consequence is that a probe-accuracy curve over these checkpoints would mix representational
change with population migration, in unknown proportion, with no way to separate them after the fact.

\begin{figure}[t]\centering
\includegraphics[width=0.62\textwidth]{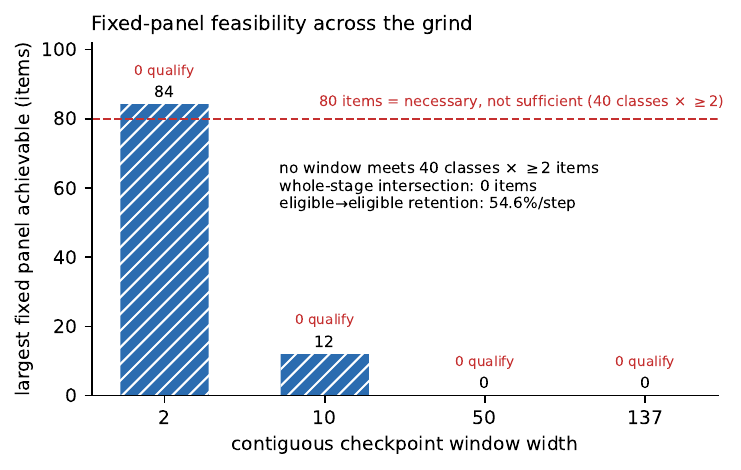}
\caption{\textbf{Why the developmental accessibility curve is not identifiable.} Largest fixed panel
of probe-eligible items achievable within a contiguous checkpoint window, by window width, against
the registered requirement of forty classes with at least two items each. No window at any width satisfies it, and the whole-stage intersection is empty. Item count alone
is necessary but not sufficient: the widest achievable panel reaches $84$ items yet still covers too
few classes, so zero windows qualify at any width. Items leave the eligible population by
improving, so eligibility is coupled to the measured event.}
\label{fig:evaluability}
\end{figure}

We established this mechanically before any developmental capture, by asking whether a fixed panel
--- the same items, in all forty classes, present across a contiguous window of checkpoints --- exists
at any window width. It does not. At width $2$, the largest achievable panel covered $84$ items and
\emph{zero} of $136$ windows met the requirement of forty classes with at least two items each; at
width $10$, zero of $128$; at width $50$, zero of $88$; at the whole-stage width, zero of $1$. The
whole-stage intersection contains no items at all. Eligible-to-eligible retention is $54.6\%$ per
step, and $8.9\%$ of items per step exit specifically by arriving. Figure~\ref{fig:evaluability}
summarises this.

Five rescues were considered and refused in advance: smoothing the curve, shortening the window,
moving the probe offsets, relaxing the support requirement, and defining a post-hoc onset statistic.
Each would have produced a number. None would have produced the registered estimand.

The result is \textsc{not cleanly evaluable}. It is not a null result, and it is not evidence that
accessibility fails to develop. It is a statement that this design cannot measure that question.

Two features make this trustworthy rather than convenient. The inferential object was fixed as
\emph{none exists} before the symbolic result of Section~\ref{sec:act2} was known, so no developmental
claim could be authored in light of an outcome. And the finding prevented rather than excused work: the
whole-stage capture never occurred, so the activations that would have supported a plausible and
invalid curve were never collected.

The general lesson is worth stating plainly, because the design that fails here is a natural one:
a checkpoint-wise probe-accuracy curve cannot by itself identify representational development when
probe eligibility changes as a function of the behaviour being tracked; without a fixed cohort or an
explicit model of that selection process, representational change and population migration are
confounded.

\subsection{A quarantined execution}
An earlier realization of the verbal accessibility experiment produced a complete result --- null
distribution, $p$-value, per-class breakdown --- while an explicit halt was in force. The cause was a
process-identity error in which the verification step was incapable of establishing the proposition it
was being used to establish; details are in the companion repository.

That population's confirmatory standing was withdrawn permanently, and the experiment was rebuilt on
an independent population, which is the result reported in Section~\ref{sec:act2}. The later positive
result does not rehabilitate the quarantined one. We report this because the alternative --- retaining
an attractive result whose execution conditions were violated --- is precisely the failure mode that
preregistration exists to prevent.

\subsection{An earlier branch that correctly returned nothing}
A separate earlier analysis returned \textsc{not evaluable} for a different reason again: the frozen
holdout geometry did not supply the support the planned statistic required. A positive control
established that the decision procedure could fire when its conditions were met, so the outcome was a
statement about support, not a failure of the instrument.

These three are distinct and should not be merged. Section~\ref{sec:act3} concerns an estimand that
does not exist under longitudinal eligibility; the quarantine concerns execution integrity with the
estimand intact; this branch concerns a population that could not support a well-defined statistic.
Only the first is a claim about identifiability.

\subsection{What richer measurement still cannot tell us}
Behaviour can miss internal accessibility. Accessibility can precede behavioural expression.
Longitudinal probing can itself become non-identifiable. And decoder success does not establish that
the decoded information is causally used. Observation, however refined, does not identify the
computation that training acquired.

\section{Related work}
\label{sec:related}

\subsection{Development across training}
That behavioural emergence can conceal gradual internal change is established: reverse-engineering the
grokking transition reveals continuous circuit formation beneath a discontinuous behavioural
curve~\citep{nanda2023progress}, and circuit-level analyses track how mechanisms appear and stabilise
across training and scale~\citep{tigges2024circuits}.

Longitudinal probing itself predates the present work. Liu et al.~\cite{liu2021probing} apply a common probe
suite across RoBERTa pretraining checkpoints to ask when linguistic, factual, commonsense and
reasoning information becomes accessible, reporting that linguistic knowledge is acquired quickly and
stably while reasoning abilities are not stably acquired. Our concern is different: when eligibility
for the probe changes as a function of the behavioural event being tracked, checkpoint-wise
accessibility can mix representational development with population migration.

Two recent neighbours are closer still. Jin et al.~\cite{jin2026finalcheckpoints} apply counterfactual edits and
activation patches at successive training checkpoints and find that latent-reasoning faithfulness
depends on training stage, arguing directly that final checkpoints are insufficient; their setting is a
different latent-reasoning architecture and task, and does not involve the recurrent-depth surface
comparison studied here. Ye et al.~\cite{ye2025implicit} train transformers from scratch in a controlled
symbolic multi-hop setting and identify a staged developmental trajectory, using cross-query semantic
patching to locate reusable intermediate representations; their question is mechanism and development
within that setting, where ours separates behavioural trajectories, inference-time answer
accessibility, and a longitudinal evaluability limit.

We therefore take the insufficiency of final checkpoints as established rather than as our
contribution. What we add is the separation of three observables --- behavioural competence,
inference-time accessibility, and training-time development --- and, in Section~\ref{sec:act3}, an
\emph{evaluability} result: a longitudinal internal measurement that cannot be identified because the
population it measures is defined by the behavioural event under study. That failure mode is a property
of the estimand rather than of the probe.

\subsection{Latent reasoning, recurrence, and what a probe licenses}
The architecture studied here iterates a shared block to scale computation at inference rather than by
emitting tokens~\citep{geiping2025recurrentdepth}, which is why the training-time and inference-time
axes must be tracked separately throughout this paper.

Two studies probe that class of model directly. Lu et al.~\cite{lu2025latentcot} examine a depth-recurrent
transformer on arithmetic with logit-lens and coda-lens decoders, find limited evidence of interpretable
latent chain-of-thought, and report that the interpretability of hidden states depends heavily on both
layer index and decoding method --- a direct caution that what is linearly readable is partly a property
of the instrument. Blayney et al.~\cite{blayney2026looped} analyse looped reasoning models mechanistically, showing
that layers in the cycle converge to distinct fixed points so that the recurrent block follows a
consistent cyclic trajectory, with attention-head behaviour stabilising as those points are reached.
Both characterise internal dynamics at inference; neither studies the developmental acquisition event,
which is where our question sits.

Our probe results are deliberately reported as accessibility rather than use, and that restraint has a
literature behind it. Ravichander et al.~\cite{ravichander2021probing} show directly that models encode properties
recoverable by a probe even when those properties are not needed for the trained task, including under
synthetic noise --- probe-decodable information need not be information the model uses --- and
Belinkov~\cite{belinkov2022probing} reviews the broader methodological limits of probing classifiers and the
difficulty of moving from decodability to causal claims. This is the reason Sections~\ref{sec:act2}
and~\ref{sec:discussion} treat linear accessibility as strictly weaker evidence than causal use, and why
the paper ends by asking for intervention rather than a better decoder.

\subsection{Representation, order, and compositional learnability}
That surface and representational choices change compositional performance is likewise established.
CLUTRR reports a substantial gap between NLU systems reasoning over natural-language stories and a
graph model given direct symbolic access to the underlying relations~\citep{sinha2019clutrr}. Because
the access pathway and model class both differ there, it does not isolate acquisition cost under
controlled surface rewrites under matched architecture and
training procedure, which is the comparison our Act~1
makes. ReCOGS shows that semantically incidental details of a target representation can dominate
measured difficulty, so that apparent model failures trace to the encoding rather than the
competence~\citep{wu2023recogs}; that result concerns its own representation and output setting rather
than an input-surface rewrite. Ramesh et al.~\cite{ramesh2024compositional} show that the order in which
compositions appear in training biases which combinations a transformer can subsequently compose, and
Lee et al.~\cite{lee2023arithmetic} show that data format and representation substantially change the sample
efficiency with which small transformers learn arithmetic. Sato et al.~\cite{sato2025orders} go further and treat
generation order explicitly as a learnability variable, searching for orders that make otherwise hard
sequential tasks trainable and diagnosing them from early-training dynamics. Training-time order is
therefore an actively studied variable, not an untouched one.

Our Act~1 result should therefore not be read as the claim that representation affects learning, which
is not in dispute. It is a measurement of the \emph{developmental cost structure} that a surface
change produces under a matched world, architecture, curriculum and criterion --- $13{,}055$ against
$70$ logical epochs at one depth, a $186.5\times$ contrast --- and of trajectories that differ in
shape rather than merely in length. We also do not decompose our surfaces into their constituent
factors: the verbal and symbolic surfaces here differ in vocabulary, grammatical structure, relation
order and sequence length simultaneously, and isolating those factors requires arms this paper does not
contain. Where Sato et al.~\cite{sato2025orders} and Ramesh et al.~\cite{ramesh2024compositional} vary order deliberately, we
vary a bundle and report its cost.

Taken together, prior work establishes that behavioural emergence can conceal gradual internal change,
that probes can reveal information not directly expressed in behaviour, and that representation and
computation order can strongly affect compositional learnability. We study these issues jointly in a
controlled recurrent-depth relational setting, separating behavioural development from inference-time
answer accessibility and showing a specific identification failure that arises when the population
eligible for longitudinal probing changes as a function of the behavioural event being measured.

\section{Discussion}
\label{sec:discussion}

\subsection{Three observables, and a fourth that is still open}
This paper separates three quantities that are often used interchangeably. \emph{Behavioural
competence} is what the model produces. \emph{Internal linear accessibility} is what a decoder can
recover from its state. \emph{Training-time development} is when either of those changes across
checkpoints. They are related and not interchangeable, and each of our three acts shows a way they can
come apart.

A fourth quantity --- the \emph{causal computation} the model performs --- is not measured here, and
the first three do not determine it.

\subsection{Surface representation changes developmental cost}
The same class of relational task, matched in world, architecture, curriculum and criterion, exhibits
radically different acquisition trajectories under two surfaces. We deliberately do not attribute that
difference to any specific property of the surfaces --- vocabulary, grammatical structure, relation
order, sequence length --- because this experiment varies them together. Decomposing that bundle
requires arms in which those factors are isolated, which is downstream of this paper. Nor do we claim
the two realizations learn the same algorithm; nothing here tests that.

\subsection{Implications for developmental interpretability}
Final checkpoints are insufficient: they cannot distinguish a long grind from a short race that ended
in the same place. Behavioural thresholds conceal route structure. Probes can reveal information that
behaviour does not express. But probe \emph{curves} can be invalid when population membership depends
on the behaviour under study, and that invalidity is not visible in the curve itself --- it has to be
established from the design in advance. Evaluability is therefore a property to be proved before
measurement, not diagnosed after. An instrument that declines to emit a number, having shown that the
number would not mean what it appears to mean, has succeeded.

\subsection{From observation to mechanism}
The next question is not whether a better decoder can read the state. It is whether the state can be
manipulated. Concretely: identify the oracle-defined semantic content of an intermediate state,
intervene on it, and test whether a recipient computation combines the transplanted intermediate with
its own untouched remainder in the way the semantics predict --- and whether a common transition law
governs that combination across entities and depths. We report none of these results here.

The observational programme has established where measurement is informative and where it is
insufficient. The next question is not whether another decoder can read the state, but whether
manipulating that state predictably changes the computation.

\section{Limitations}
\label{sec:limitations}
These are scope statements, not caveats offered in mitigation.

The results concern one closed synthetic world and one recurrent-depth architecture class, and do not
automatically generalise to frontier language models. The answer universe is frozen at forty classes;
we test neither unseen-answer nor unseen-head generalisation. The probe effects are weak and strongly
heterogeneous across classes --- $21/40$ and $16/40$ classes contribute, and the thinnest classes
contribute nothing --- so the pooled figures are not uniform properties of the class space. The two
surfaces are separate trained realizations, not one model observed twice. No registered
cross-population effect-size comparison between surfaces was performed, so the surfaces are compared
descriptively only. Linear accessibility does not establish a completed representation, and no causal
use, shared representation across surfaces, or shared transition law is established. No cross-world
confirmation exists. Verbal $k=5$ is right-censored at the manuscript snapshot: at the final evaluated filename epoch
$20{,}399$ its in-stage cost was at least $6{,}400$ logical epochs with a maximum observed accuracy of
$0.9133$ and no crossing of the $0.95$ criterion. And the developmental accessibility trajectory is
not cleanly evaluable in this design, so the question it was meant to answer remains open.

\section{Conclusion}
Behavioural thresholds hide developmental structure: realizations matched in world, architecture, curriculum and acquisition criterion but trained on
different surfaces required $70$ and $13{,}055$ logical epochs for the same depth of competence, and a
$13{,}055$-epoch grind was followed by an $8$-epoch race. Weak answer information is linearly
accessible before behavioural arrival, and this survives a change of surface, appearing at two
separately preregistered positions. And richer internal observation can itself encounter an
identification failure, when the population a probe measures is defined by the behaviour being
studied.

We can observe more than behaviour reveals, but observation still does not identify what computation
training acquired.

\section*{Data and code availability}
Instrument sources, registered specifications, per-result receipts with cryptographic digests, the
population and support proofs underlying Section~\ref{sec:act3}, and the figure-generation scripts for
every panel in this paper are available in the companion repository at
\url{https://github.com/primecalibre-research/ltg-replication-receipts}.
Executed registrations and receipts supporting reported claims are released with the paper.
Prospectively registered but unexecuted experiments remain sealed until execution or formal
retirement to preserve prospective integrity.

\bibliographystyle{plain}
\bibliography{references}

\begin{thebibliography}{10}

\bibitem{belinkov2022probing}
Yonatan Belinkov.
\newblock Probing classifiers: Promises, shortcomings, and advances.
\newblock {\em Computational Linguistics}, 48(1):207--219, 2022.

\bibitem{blayney2026looped}
Hugh Blayney, {\'A}lvaro Arroyo, Johan Obando-Ceron, Pablo~Samuel Castro, Aaron
  Courville, Michael~M. Bronstein, and Xiaowen Dong.
\newblock A mechanistic analysis of looped reasoning language models.
\newblock 2026.

\bibitem{geiping2025recurrentdepth}
Jonas Geiping, Sean McLeish, Neel Jain, John Kirchenbauer, Siddharth Singh,
  Brian~R. Bartoldson, Bhavya Kailkhura, Abhinav Bhatele, and Tom Goldstein.
\newblock Scaling up test-time compute with latent reasoning: A recurrent depth
  approach.
\newblock 2025.
\newblock NeurIPS 2025 spotlight.

\bibitem{jin2026finalcheckpoints}
Hengyu Jin, Shu Yang, and Di~Wang.
\newblock Final checkpoints are not enough: Analyzing latent reasoning
  faithfulness along training trajectories.
\newblock 2026.

\bibitem{lammuir2026ignition}
Simon Lam-Muir.
\newblock The ignition is real, and it lives at the readout: Latent
  composition, difficulty-clocked ignition, and the interface-constituted
  commit in a recurrent-depth reasoner.
\newblock 2026.

\bibitem{lee2023arithmetic}
Nayoung Lee, Kartik Sreenivasan, Jason~D. Lee, Kangwook Lee, and Dimitris
  Papailiopoulos.
\newblock Teaching arithmetic to small transformers.
\newblock 2023.

\bibitem{liu2021probing}
Leo~Z. Liu, Yizhong Wang, Jungo Kasai, Hannaneh Hajishirzi, and Noah~A. Smith.
\newblock Probing across time: What does {RoBERTa} know and when?
\newblock In {\em Findings of the Association for Computational Linguistics:
  EMNLP 2021}, 2021.

\bibitem{lu2025latentcot}
Wenquan Lu, Yuechuan Yang, Kyle Lee, Yanshu Li, and Enqi Liu.
\newblock Latent chain-of-thought? decoding the depth-recurrent transformer.
\newblock In {\em First Workshop on the Application of LLM Explainability to
  Reasoning and Planning, Conference on Language Modeling (COLM)}, 2025.

\bibitem{nanda2023progress}
Neel Nanda, Lawrence Chan, Tom Lieberum, Jess Smith, and Jacob Steinhardt.
\newblock Progress measures for grokking via mechanistic interpretability.
\newblock In {\em International Conference on Learning Representations (ICLR)},
  2023.

\bibitem{ramesh2024compositional}
Rahul Ramesh, Ekdeep~Singh Lubana, Mikail Khona, Robert~P. Dick, and Hidenori
  Tanaka.
\newblock Compositional capabilities of autoregressive transformers: A study on
  synthetic, interpretable tasks.
\newblock In {\em Proceedings of the 41st International Conference on Machine
  Learning (ICML)}, 2024.

\bibitem{ravichander2021probing}
Abhilasha Ravichander, Yonatan Belinkov, and Eduard Hovy.
\newblock Probing the probing paradigm: Does probing accuracy entail task
  relevance?
\newblock In {\em Proceedings of the 16th Conference of the European Chapter of
  the Association for Computational Linguistics (EACL)}, 2021.

\bibitem{sato2025orders}
Yuta Sato, Kazuhiko Kawamoto, and Hiroshi Kera.
\newblock Discovering learning-friendly generation orders for sequential
  computation.
\newblock 2025.
\newblock arXiv preprint; v1 June 2025, revised May 2026.

\bibitem{sinha2019clutrr}
Koustuv Sinha, Shagun Sodhani, Jin Dong, Joelle Pineau, and William~L.
  Hamilton.
\newblock {CLUTRR}: A diagnostic benchmark for inductive reasoning from text.
\newblock In {\em Proceedings of the 2019 Conference on Empirical Methods in
  Natural Language Processing and the 9th International Joint Conference on
  Natural Language Processing (EMNLP-IJCNLP)}, 2019.

\bibitem{tigges2024circuits}
Curt Tigges, Michael Hanna, Qinan Yu, and Stella Biderman.
\newblock {LLM} circuit analyses are consistent across training and scale.
\newblock In {\em Advances in Neural Information Processing Systems (NeurIPS)},
  2024.

\bibitem{wu2023recogs}
Zhengxuan Wu, Christopher~D. Manning, and Christopher Potts.
\newblock {ReCOGS}: How incidental details of a logical form overshadow an
  evaluation of semantic interpretation.
\newblock {\em Transactions of the Association for Computational Linguistics},
  11:1719--1733, 2023.

\bibitem{ye2025implicit}
Jiaran Ye, Zijun Yao, Zhidian Huang, Liangming Pan, Jinxin Liu, Yushi Bai, Amy
  Xin, Weichuan Liu, Xiaoyin Che, Lei Hou, and Juanzi Li.
\newblock How do transformers learn implicit reasoning?
\newblock In {\em Advances in Neural Information Processing Systems (NeurIPS)},
  2025.
\newblock Spotlight.

\end{thebibliography}

\end{document}